\documentclass[runningheads]{llncs}
\usepackage{amsfonts}
\usepackage{ctable}
\usepackage{tabularx}
\usepackage{graphicx}
\usepackage{hyperref}
\hypersetup{
    colorlinks=true,
    linkcolor=blue,
    filecolor=magenta,      
    urlcolor=magenta
}

\newcolumntype{Z}{>{\centering\arraybackslash}X}
\newcommand{\bftab}{\fontseries{b}\selectfont}

\begin{document}

\title{Autism spectrum disorder classification based on interpersonal neural synchrony: Can classification be improved by dyadic neural biomarkers using unsupervised graph representation learning?}
\titlerunning{ASD classification based on interbrain networks}
%
\author{
\index{Last Name, First Name}
\index{Last Name, First Name} 
\index{Last Name, First Name} 
\index{Last Name, First Name} 
\index{Last Name, First Name} 
Christian Gerloff  \inst{1, 2, 3} \and
Kerstin Konrad \inst{1,2} \and
Jana Kruppa \inst{1, 4} \and
Martin Schulte-Rüther \thanks{Shared last authorship.} \inst{4, 5} \and
Vanessa Reindl* \inst{1, 2, 6}}
\authorrunning{C. Gerloff et al.}
%
\institute{JARA-Brain Institute II, Molecular Neuroscience and Neuroimaging, RWTH Aachen \& Research Centre Juelich, Aachen, Germany \and
Child Neuropsychology Section, Department of Child and Adolescent Psychiatry, Psychosomatics and Psychotherapy, Medical Faculty, RWTH Aachen University, Aachen, Germany \and
Chair II of Mathematics, Faculty of Mathematics, Computer Science and Natural Sciences, RWTH Aachen University, Aachen, Germany \and
Department of Child and Adolescent Psychiatry and Psychotherapy, University Medical Center Göttingen, Göttingen, Germany
 \and
Leibniz ScienceCampus Primate Cognition, Göttingen, Germany \and
Psychology, School of Social Sciences, Nanyang Technological University, Singapore, Singapore \\
}
\maketitle              
\begin{abstract} Research in machine learning for autism spectrum disorder (ASD) classification bears the promise to improve clinical diagnoses. However, recent studies in clinical imaging have shown the limited generalization of biomarkers across and beyond benchmark datasets. Despite increasing model complexity and sample size in neuroimaging, the classification performance of ASD remains far away from clinical application. This raises the question of how we can overcome these barriers to develop early biomarkers for ASD. One approach might be to rethink how we operationalize the theoretical basis of this disease in machine learning models. Here we introduced unsupervised graph representations that explicitly map the neural mechanisms of a core aspect of ASD, deficits in dyadic social interaction, as assessed by dual brain recordings, termed hyperscanning, and evaluated their predictive performance. The proposed method differs from existing approaches in that it is more suitable to capture social interaction deficits on a neural level and is applicable to young children and infants. First results from functional near-infrared spectroscopy data indicate potential predictive capacities of a task-agnostic, interpretable graph representation. This first effort to leverage interaction-related deficits on neural level to classify ASD may stimulate new approaches and methods to enhance existing models to achieve developmental ASD biomarkers in the future.

\keywords{Unsupervised graph representation learning  \and ASD \and interbrain networks \and interpersonal neural synchrony \and hyperscanning \and functional near-infrared spectroscopy}
\end{abstract}
\section{Introduction} \label{Introduction} 
\raggedbottom
Autism spectrum disorder (ASD) is a neurodevelopmental disorder that affects approximately 1\% of the population \cite{Zeidan2022} and is characterized by impairments in reciprocal social interaction, communication, and repetitive stereotypic behavior \cite{APA2013}. Of these symptoms, deficits in social interaction are often considered as most central to the disorder \cite{Scott2010}. Since ASD is associated with a high global burden of disease \cite{Baxter2015}, early detection and intervention are of utmost importance for optimal long-term outcomes \cite{Dawson2010}. Diagnosing ASD is currently exclusively based on behavioral observation and anamnestic information. Especially at early ages, it is a challenging, time-consuming task, requiring a high level of clinical expertise and experience.

Brain imaging may provide a complementary source of information. Over the last decades, a vast body of research has documented anatomical and functional brain differences between individuals with ASD and healthy controls (e.g., \cite{Ecker2015}), however, despite increasing model complexity they failed to reveal consistent neural biomarkers. Current machine learning models predict ASD diagnosis based on functional magnetic resonance imaging data from large datasets (n$>$2000) with an area under the curve (AUC) of $\sim$0.80 \cite{Traut2022}. However, even when information leakage is avoided by adequate cross-validation (CV) procedures (e.g., \cite{Traut2022,Hosseini2020}), the predictive accuracy typically decreases in medical imaging applications when validating with private hold-out sets (see \cite{Pooch2020,Varoquaux2022}, \cite{Traut2022}: from $AUC=0.80$ to $AUC=0.72$). Increasing sample sizes may improve this situation, however, this is often not feasible with clinical data and recent results using the largest database currently available for ASD (ABIDE) suggest an asymptotic behavior of $AUC$, with $AUC=0.83$ for 10,000 ASD subjects \cite{Traut2022}. While an $AUC=0.83$ is promising, this is still far away from classification accuracies (AUC of up to 0.93) when applying machine learning methods to behavioral data obtained from a social interactive situation, such as ADOS \cite{Ruther2021}. One approach to overcome these roadblocks might be to rethink how we operationalize the theoretical basis of disease in machine learning models.

Although ASD is characterized by social difficulties during interaction with others \cite{Bolis2018}, MRI data used for ASD classification has typically been acquired without any social interactive context. Participants lay still and alone in an MRI scanner during resting state and structural scans (e.g., \cite{Traut2022}) and only a few studies used simple social tasks, such as passively viewing social scenarios (e.g., \cite{Chanel2016}). We suggest that a more ecologically valid neurobiological measure of social interaction could be derived from dyadic setups that simultaneously assess the brains of interacting subjects. Such “hyperscanning” settings typically use less intrusive imaging techniques, such as functional near-infrared spectroscopy (fNIRS), that are more tolerable to young children. Previous studies using a variety of tasks \cite{Babiloni2014} have demonstrated that statistical dependencies of brain activity emerge across individuals, suggesting interpersonal neural synchrony (INS). Initial results in ASD suggest reduced INS during joint attention, communication and cooperative coordination tasks (\cite{Tanabe2012,Camacho2021,Wang2020}, but see \cite{Kruppa2021} for contradictory findings). Considering that the clinical diagnosis of ASD is crucially dependent on analysing interactive, reciprocal behavior, it appears particularly promising to employ such interbrain measures for the classification of ASD.

Recently, we formalized INS using bipartite graphs, with brain regions of the two participants as two sets of nodes and inter-brain connections as edges \cite{Gerloff2022}. We proposed to accommodate the non-Euclidean structures using network embeddings to predict experimental conditions from inter-brain networks. Generally, to discriminate among graphs that belong to different classes or clinical groups requires a graph representation that can either be part of an end-to-end model, such as graph neural networks (e.g.,\cite{Kipf2016}), or of an encoder, e.g., Graph2Vec \cite{Narayanan2017}. The former is typically trained in a supervised or semi-supervised fashion to learn class associations directly. The latter requires an intermediate step, i.e. unsupervised training to derive a task-agnostic representation which can subsequently be used for classification using
a range of available algorithms. This allows to compare the predictive capacities of connectivity estimators vs. graph representations based on the same classifier. Additionally, task-related intrabrain data indicate that classifiers might benefit from lower-vector representations when sample sizes are low \cite{Brodersen2014}.

To summarize, we aim to contribute to the field of machine learning for clinical imaging by exploring, for the first time, unsupervised graph representation learning and its potential to contribute to the classification of ASD using a hyperscanning dataset \cite{Kruppa2021}. Since such datasets are rare and typically small, this should be considered as a first step to demonstrate feasibility and to encourage 1) application of these methods to other tasks and 2) consider the collection of larger hyperscanning cohorts to further explore this approach. Specifically, we
aim to contribute by:

\begin{itemize}
\renewcommand{\labelitemi}{\--}
\item {\bfseries Operationalization of theoretical basis.} We propose a new methodological approach that aims to encode a core aspect of ASD (i.e. deficits in social interaction) at the neural level by applying network embeddings on graph representations of interpersonal neural synchrony.
\item {\bfseries Predictive validity.} We assessed the predictive validity to classify ASD dyads based on graph representations of interbrain networks. We examined how the employed connectivity estimator and different types of network embeddings influence classification performance.
\item {\bfseries Future contribution to early life biomarkers and beyond.} The proposed method differs from existing approaches in that it i) allows to capture social interaction deficits on a neural level and ii) is applicable to young children and infants. Moreover, we present a task-agnostic graph representation of interbrain synchrony that is applicable for ASD classification and beyond, e.g., using statistical models for inferential inquiries.
\end{itemize}

\section{Methods}
\subsection{Problem Formulation} 
We formalize the problem of whether graph representations of INS can classify participants with or without ASD diagnosis. 

First, given a set of graphs denoted by ${X=\left(G_1,\ldots,G_n\right)}$ we intend to train an encoder $\phi$ parameterized by $\Theta^E$. The encoder learns unsupervised from similarities in graph space a matrix representation of all graphs $Z \in R^{|\mathbb{X}| \times \delta}$ with $Z_i \in R^{\delta}$ representing the graph $G_i \in X $. The size $\delta$ of the network embedding is either parameterized via $\Theta^E$ or given by the encoder.

Second, the predictive capacities of various $\phi$ are evaluated in an inductive learning regime by a classifier $\hat{C}:\mathcal{Z}\mapsto Y$ parameterized by $\Theta^C$. The labels denoted by $Y=\left(y_{1}, \ldots, y_{n}\right)$ represent graphs with ASD ($y_i=1$) or without ASD ($y_i=0$). Finally, the performance metric $e\left(\hat{y};y\right)$ of each $\phi$ is assessed.

\subsection{Functional connectivity} \label{FC}
Functional connectivity estimators (FC) quantify the statistical relationship between two neural signals. They can be used for further network construction (see definition \ref{definition}) or serve as features for ASD classification (e.g., in interbrain studies \cite{Traut2022}). Here, we evaluate whether graph representations carry further beneficial information (e.g., topological properties) than the pure connectivity estimator. As connectivity estimators vary in the captured dynamics between signals, we calculated, based on the continuous wavelet transform, the following estimators to systematically account for different aspects of the dynamics: wavelet coherence (WCO), phase-locking value (PLV), Shannon entropy (Entropy). WCO captures mostly concurrent synchronization between two brain signals of two participants in time-frequency space. While WCO also accounts for amplitude differences, PLV considers only phase synchronization. In contrast to WCO and PLV, Entropy captures non-linear, delayed forms of synchronization.

\subsection{Graph representation learning}
While FCs describe a pairwise association between two brain regions, the estimator itself does not capture the multiple interdependencies and topological properties of the brain which has been shown to be organized and function as a network. A system formulation that accounts for these network characteristics and encodes social interaction, a core aspect of ASD, at neural level are interbrain networks \cite{Gerloff2022}.

\begin{definition}\label{definition} In accordance with \cite{Gerloff2022}, interbrain networks can be specified by a bipartite graph ${G=\left(V_1\cup V_2,E\right)}$ where $V_1$ and $V_2$ denote brain regions of participant 1 and participant 2, respectively. $E\subseteq V_1\times V_2$ represents the edges with the corresponding weights $W$, defined by the specific connectivity estimator.
\end{definition}

Based on three distinct FCs and the subsequently derived interbrain networks, we assessed the predictive capacities of state-of-the-art graph representation learning. In the following, we describe what sets each encoder apart.

\subsubsection{NMF-based interbrain network embedding (NMF-IBNE)}
 provides a lower-vector representation that encodes proximity of the topological properties of a bipartite graph \cite{Gerloff2022}. This embedding does not assume a connected bipartite graph and can operate together with graph reduction procedures stratifying for non-interaction related connectivity. It leverages substructures based on a priori specified graph properties of interest (here, nodal density). Importantly, its basis matrix enables a direct interpretation of the contribution of each brain region due to the non-negative constraints of the NMF.

\subsubsection{Local Degree Profile (LDP)} encodes the graph structure based on the nodal degree of a node and its neighbors \cite{Cai2018}. For this purpose, the encoder maps the nodal degree, its mean, minimum, maximum, and standard deviation of the first neighbors via a histogram or empirical distribution function.

\subsubsection{Graph2Vec} employs the neural network architecture skip-gram in graph space \cite{Narayanan2017}. Conceptually, given $X$ and a sequence of sampled subgraphs from different nodes, the algorithm minimizes log-likelihood of the rooted subgraph corresponding to a specific $G$. Each subgraph is derived via the Weisfeiler-Lehman relabeling process around each node.

\subsubsection{GL2Vec} aims to adjust Graph2Vec for an edge-centric case \cite{Chen2019}. In the same spirit, it minimizes the likelihood of the rooted subgraph but instead of operating directly on $G$, it transforms each graph into a line graph. Thereby, it extracts edge-centric substructures and enhances the integration of edge properties, which sets it apart from the other approaches.

\subsubsection{Diffusion-wavelet-based node feature characterization (DWC)} describes an algorithm that assesses the topological similarities using a diffusion wavelet and node features \cite{Wang2021}. The eigenvalues of the Laplacian matrix describe the temporal frequencies of a signal on $G$. Coefficients derived from a wavelet kernel represent the energy transferred between nodes, whereas nodes with similar energy patterns have similar structural roles.

\subsubsection{Geometric Scattering (Scattering)} applies the invariant scattering transform on graphs \cite{Gao2019}. Like DWC, the algorithm encodes topological similarities via diffusion wavelets but on the normalized Laplacian to consider nodal degree.

\subsubsection{Feather} performs random walks on the normalized adjacency matrix \cite{Rozemberczki2020}. Assuming that the correlation between node properties is related to the role similarity, it pools the r-scaled weighted characteristic from the adjacency matrix.

\subsection{Classifier}
We employed two common classifiers for ASD classification, specifically, L2 regularized logistic regression and support-vector-machines (e.g., \cite{Traut2022}).

\subsection{Hyperparameter optimization}
Let $\Theta= \left\{\Theta^E, \Theta^C\right\}$ denote the model hyperparameters, we aim to select a model parametrization by splitting each training set $\mathcal{D}_{train}$ into $k_{inner}=3$  training and distinct test sets. Hyperparameter optimization was performed via a Gaussian Process based on this cross-validation setting. The area under the curve of the receiver operating characteristic (ROC-AUC) was used as the evaluation score to estimate $\hat{\Theta}={arg\ min}_\Theta L\left(C_\Theta;\mathcal{D}_{train}\right)$.

\section{Experiment}
\subsection{Dataset}
To examine the capacities of the graph representations for classifying ASD, we used the hyperscanning dataset provided by \cite{Kruppa2021}. The cohort consists of 18 children and adolescents diagnosed with ASD and 18 typically developed children and adolescents, matched in age (8 and 18 years) and gender. Each child performed a cooperative and competitive computer game with the parent and an adult stranger in two task blocks. During the game, the prefrontal brain activities were measured concurrently using fNIRS. fNIRS is an optical imaging technique that measures neural activity through concentration changes in oxygenated (HbO) and deoxygenated (HbR) hemoglobin. The brain's metabolic demand for oxygen and glucose increases in active brain areas. Thereby, the concentration of HbO increases and HbR decreases. HbO and HbR were preprocessed consistently with \cite{Kruppa2021}. Overall, this allows the construction of 136 ASD-related and 144 healthy control-related graphs for HbO and HbR each (see section \ref{FC}).

\subsection{Evaluation regimes}\label{Evaluation}
We examine the capacities to discriminate between graphs according to ASD status by performing i) CV and ii) a cross-chromophore test (CCT; see \cite{Gerloff2022}). While HbO is most commonly analyzed in fNIRS studies, HbO was used in (i) and HbR served as the test set in (ii). Importantly, in (i) and (ii) unsupervised representation learning was evaluated in an inductive learning regime in which we do not expose $\phi$ to test data during training. In (i), we employed a nested-stratified CV to ensure a generalizable evaluation and avoid information leakage during hyperparameter optimization. To this end, $X$ was randomly partitioned into $k_{out}=5$ mutually exclusive subsets, where we strictly ensure that all data from one dyad is in one subset and that the proportion of ASD to healthy control dyads is preserved to ensure generalization across dyads. HbO and HbR rely on the same neurovascular coupling mechanism but have inverse characteristics. In (ii), training was performed in an isolated fashion on HbO to assess the test performance on HbR-based networks. This kind of out-of-distribution test is a unique opportunity of fNIRS to test predictive capacities across chromophores. For both evaluation regimes (i, ii), performance was quantified by ROC-AUC as an established performance metric in ASD classification \cite{Traut2022}. For (i), mean and standard deviation across all folds were reported. In clinical settings, an evaluation of the performance variance might be particularly important (see also \cite{Bouthillier2021}). Thus, we employed a Bayesian correlated t-test \cite{Benavoli2017} accounting for the correlation of overlapping training sets. Further, by randomly shuffling $Y$, we tested the robustness of the results against training on randomly labeled data.

\subsection{Implementation details and reproducibility}
\subsubsection{Code,  parameter, and metric versioning.} Mlflow served as a tracking and versioning environment. Parameters and performance metrics were stored in a non-relational database. The versioned code of embeddings and evaluation regimes is available in the repository: \url{https://github.com/ChristianGerloff/IBN-ASD-classification}

\subsubsection{Unit tests and technical reproducibility.}
To accommodate best practices from software development, the repository includes unit tests. Dependencies were managed using Poetry. Experiments were run inside a docker container on a remote instance with 4 x 3.1 GHz Intel Xeon processors and 16 GiB memory.

\subsection{ASD classification performance}
Summarized, the rigorous and conservative evaluation demonstrated the challenges in ASD classification. Only results of NMF-IBNE based on Shannon entropy indicate potential capacities to predict ASD in this dataset. FC may be less robust as it suffered from particularly high variance in ROC-AUC on randomly labeled data. This might indicate that INS-based ASD classification could potentially benefit from graph representations.

\begin{figure}[!htb]
\includegraphics[width=\textwidth]{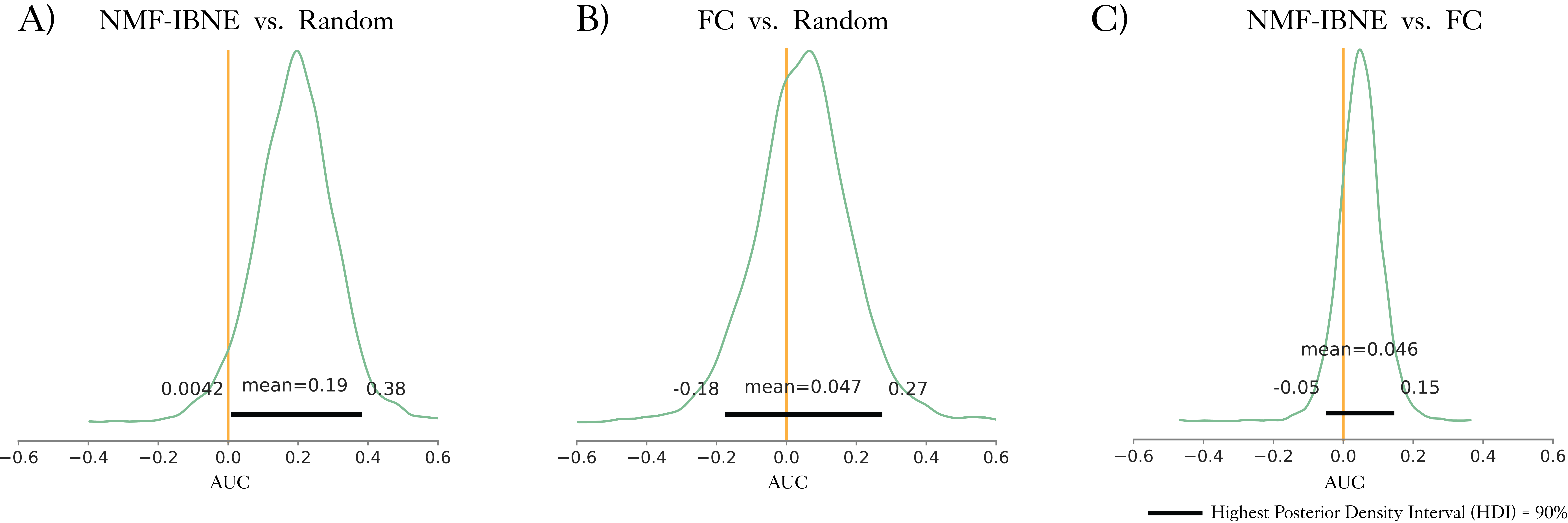}
\caption{(A) Strong evidence was found for superior AUC of NMF-IBNE compared to model performance on randomized labels. (B) FC showed high uncertainty and no evidence of predictive validity in the randomization test. (C) Posterior distribution indicated a weak tendency of increased AUC in NMF-IBNE compared to FC.} \label{fig1}
\end{figure}

Specifically, the CV results shown in table \ref{table1} suggest that concurrent and linear forms of interaction-related synchrony (WCO, PLV) did not allow to differentiate between ASD and healthy subjects,
in line with findings on population level \cite{Kruppa2021}. In contrast, across both classifiers, FC and NMF-IBNE showed AUC above chance level, indicating that delayed, nonlinear forms of synchrony (Entropy) may capture ASD-related aspects of social interaction. To verify these results, a test based on randomized labels was performed (see section \ref{Evaluation}). Models above chance level that passed this test are marked in bold in table \ref{table1}. Only NMF-IBNE performed better than training on randomly labeled data across classifiers (HDI lies right to zero, see fig\ref{fig1} A, B), speaking for the robustness of the results. However, in this small cohort, NMF-IBNE showed only a weak tendency for increased performance of NMF-IBNE compared to FC (see fig\ref{fig1}C). Importantly, in contrast to other graph representation models, NMF-IBNE yields model intrinsic interpretability that enables to study the neural basis of ASD from an inferential perspective (see also \cite{Reindl2022}).

CCT revealed that HbO and HbR embeddings are distinct. In addition to systematic differences between HbR and HbO, both are differentially affected by physiological and motion artifacts.

\begin{table}[h]
\caption{Classification performance of cross-validation and cross-chromophore test.} \label{table1}
\centering
\begin{tabularx}{\textwidth}{ccZZZZZc}
& & \multicolumn{2}{c}{ WCO } &  \multicolumn{2}{c}{ PLV } &  \multicolumn{2}{c}{ Entropy } \\
\cline { 3 - 8 }  &  & SVM &        Ridge &          SVM &        Ridge &          SVM &        Ridge \\
\specialrule{.01em}{.0em}{.0em}
CV & FC &  0.50±0.04 &  0.55±0.09 &  0.53±0.04 &  0.55±0.06 &  0.53±0.11 &  0.59±0.07 \\
    & NMF\--IBNE &  0.54±0.11 &  0.48±0.07 &  0.52±0.05 &  0.54±0.06 &  \bftab 0.60±0.05 &  \bftab 0.61±0.05 \\
    & LDP &  0.49±0.02 &  0.51±0.09 &  0.51±0.07 &  0.43±0.06 &  0.47±0.11 &  0.51±0.09 \\
    & Graph2Vec &  0.50±0.01 &  0.50±0.01 &  0.52±0.05 &  0.50±0.04 &  0.54±0.07 &  0.50±0.01 \\
    & GL2Vec &  0.53±0.04 &  0.51±0.01 &  0.50±0.00 &  0.52±0.03 &  0.49±0.09 &  0.50±0.00 \\
    & DWC &  0.48±0.08 &  0.48±0.03 &  0.49±0.02 &  0.47±0.03 &  0.46±0.12 &  0.50±0.11 \\
    & Scattering &  0.56±0.04 &  0.49±0.07 &  0.48±0.03 &  0.43±0.04 &  0.53±0.07 &  0.54±0.10 \\
    & Feather &  0.47±0.06 &  0.46±0.08 &  0.52±0.03 &  0.50±0.04 &  0.49±0.04 &  0.54±0.08 \\
\hline
CCT & FC &       0.53 &       0.56 &       0.50 &       0.51 &       0.56 &     0.61 \\
    & NMF\--IBNE &       0.46 &       0.46 &       0.56 &       0.53 &       0.50 &       0.50 \\
    & LDP &       0.52 &       0.54 &       0.49 &       0.47 &       0.50 &       0.58 \\
    & Graph2Vec &       0.49 &       0.50 &       0.49 &       0.49 &       0.47 &       0.49 \\
    & GL2Vec &       0.50 &       0.50 &       0.50 &       0.50 &       0.49 &       0.50 \\
    & DWC &       0.57 &       0.57 &       0.50 &       0.50 &       0.50 &       0.57 \\
    & Scattering &       0.57 &       0.52 &       0.49 &       0.49 &       0.56 &       0.55 \\
    & Feather &       0.50 &       0.51 &       0.51 &       0.48 &       0.57 &       0.57 \\
\hline
\end{tabularx}
\end{table}

\section{Conclusion}
Recent studies in clinical imaging and other areas of machine learning have shown that despite an increasing model complexity, generalization across benchmark datasets may be limited. Therefore, current ASD classification performance using resting-state data may not be sufficient for clinical applications, even when sample size increases. Developing methods that provide a more precise representation of neural data which is better suited to capture the specific aspects of a disorder is of utmost importance.

Here we introduced unsupervised learning of graph representations that explicitly map the neural mechanisms of a core aspect of ASD (i.e. deficits in dyadic social interaction) and combined these with classification algorithms. We employed a rigorous evaluation regime to ensure predictive validity.

These first results indicate that ASD classification based on hyperscanning data is still challenging, yet network embeddings might contribute to improve the development of biomarkers. Furthermore, the choice of the connectivity estimator appears to be important. In particular, nonlinear
structures of time-varying synchrony, e.g. captured by Shannon entropy, should
be explored in greater detail in hyperscanning. Further, the results indicate that HbO and HbR, which are differentially affected by physiological effects and artifacts, might be treated as related but distinct features in future ASD studies.
Certainly, further benchmarking is necessary in larger samples, however, our results indicate that the collection of such datasets should be pushed forward to advance the development and validation of neural biomarkers in ASD. Importantly, our proposed method is suitable for young children, an age at which behavioral diagnostic markers are less reliable but at the same time, it is an early detection and intervention that can dramatically increase long-term outcomes in ASD. Notably, this approach may move forward the scientific investigation
of ASD beyond diagnostic classification: Interpretable network embeddings such as NMF-IBNE are trained in a task-agnostic fashion, are thus applicable to many settings and allow for flexible downstream applications, e.g. to address inferential scientific inquiries such as changes after intervention.

%
%
%
\bibliographystyle{splncs04}
\bibliography{paper17.bib}

\begin{thebibliography}{10}
\providecommand{\url}[1]{\texttt{#1}}
\providecommand{\urlprefix}{URL }
\providecommand{\doi}[1]{https://doi.org/#1}

\bibitem{APA2013}
Association, A.P.: Diagnostic and statistical manual of mental disorders.
  American Psychiatric Association  \textbf{21}(5),  591--643 (2013)

\bibitem{Babiloni2014}
Babiloni, F., Astolfi, L.: Social neuroscience and hyperscanning techniques:
  past, present and future. Neuroscience \& Biobehavioral Reviews  \textbf{44},
   76--93 (2014)

\bibitem{Baxter2015}
Baxter, A.J., Brugha, T., Erskine, H.E., Scheurer, R.W., Vos, T., Scott, J.G.:
  The epidemiology and global burden of autism spectrum disorders.
  Psychological medicine  \textbf{45}(3),  601--613 (2015)

\bibitem{Benavoli2017}
Benavoli, A., Corani, G., Demšar, J., Zaffalon, M.: Time for a {Change}: a
  {Tutorial} for {Comparing} {Multiple} {Classifiers} {Through} {Bayesian}
  {Analysis}. Journal of Machine Learning Research  \textbf{18}(77),  1--36
  (2017)

\bibitem{Bolis2018}
Bolis, D., Schilbach, L.: Observing and participating in social interactions:
  action perception and action control across the autistic spectrum.
  Developmental cognitive neuroscience  \textbf{29},  168--175 (2018)

\bibitem{Bouthillier2021}
Bouthillier, X., Delaunay, P., Bronzi, M., Trofimov, A., Nichyporuk, B., Szeto,
  J., Mohammadi~Sepahvand, N., Raff, E., Madan, K., Voleti, V., Ebrahimi~Kahou,
  S., Michalski, V., Arbel, T., Pal, C., Varoquaux, G., Vincent, P.: Accounting
  for {Variance} in {Machine} {Learning} {Benchmarks}. In: Smola, A., Dimakis,
  A., Stoica, I. (eds.) Proceedings of {Machine} {Learning} and {Systems}.
  vol.~3, pp. 747--769 (2021)

\bibitem{Brodersen2014}
Brodersen, K.H., Deserno, L., Schlagenhauf, F., Lin, Z., Penny, W.D., Buhmann,
  J.M., Stephan, K.E.: Dissecting psychiatric spectrum disorders by generative
  embedding. NeuroImage  \textbf{4},  98--111 (2014)

\bibitem{Cai2018}
Cai, C., Wang, Y.: A simple yet effective baseline for non-attributed graph
  classification. arXiv preprint arXiv:1811.03508  (2018)

\bibitem{Chanel2016}
Chanel, G., Pichon, S., Conty, L., Berthoz, S., Chevallier, C., Grèzes, J.:
  Classification of autistic individuals and controls using cross-task
  characterization of {fMRI} activity. NeuroImage  \textbf{10},  78--88 (2016)

\bibitem{Chen2019}
Chen, H., Koga, H.: {GL2vec}: {Graph} {Embedding} {Enriched} by {Line} {Graphs}
  with {Edge} {Features}. Neural Information Processing  \textbf{11955},  3--14
  (2019)

\bibitem{Dawson2010}
Dawson, G., Rogers, S., Munson, J., Smith, M., Winter, J., Greenson, J.,
  Donaldson, A., Varley, J.: Randomized, controlled trial of an intervention
  for toddlers with autism: the {Early} {Start} {Denver} {Model}. Pediatrics
  \textbf{125}(1),  e17--e23 (2010)

\bibitem{Ecker2015}
Ecker, C., Bookheimer, S.Y., Murphy, D.G.: Neuroimaging in autism spectrum
  disorder: brain structure and function across the lifespan. The Lancet
  Neurology  \textbf{14}(11),  1121--1134 (2015)

\bibitem{Gao2019}
Gao, F., Wolf, G., Hirn, M.: Geometric {Scattering} for {Graph} {Data}
  {Analysis}. In: Proceedings of the 36th {International} {Conference} on
  {Machine} {Learning}. pp. 2122--2131. PMLR (May 2019), iSSN: 2640-3498

\bibitem{Gerloff2022}
Gerloff, C., Konrad, K., Bzdok, D., Büsing, C., Reindl, V.: Interacting brains
  revisited: {A} cross‐brain network neuroscience perspective. Human Brain
  Mapping  \textbf{43}(14),  4458--4474 (2022)

\bibitem{Hosseini2020}
Hosseini, M., Powell, M., Collins, J., Callahan-Flintoft, C., Jones, W.,
  Bowman, H., Wyble, B.: I tried a bunch of things: {The} dangers of unexpected
  overfitting in classification of brain data. Neuroscience \& Biobehavioral
  Reviews  \textbf{119},  456--467 (2020)

\bibitem{Kipf2016}
Kipf, T.N., Welling, M.: Semi-supervised classification with graph
  convolutional networks. arXiv preprint arXiv:1609.02907  (2016)

\bibitem{Kruppa2021}
Kruppa, J.A., Reindl, V., Gerloff, C., Oberwelland~Weiss, E., Prinz, J.,
  Herpertz-Dahlmann, B., Konrad, K., Schulte-Rüther, M.: Brain and motor
  synchrony in children and adolescents with {ASD}—a {fNIRS} hyperscanning
  study. Social cognitive and affective neuroscience  \textbf{16}(1-2),
  103--116 (2021)

\bibitem{Narayanan2017}
Narayanan, A., Chandramohan, M., Venkatesan, R., Chen, L., Liu, Y., Jaiswal,
  S.: graph2vec: {Learning} {Distributed} {Representations} of {Graphs}. arXiv
  preprint arXiv:1707.05005  (2017)

\bibitem{Pooch2020}
Pooch, E.H., Ballester, P., Barros, R.C.: Can we trust deep learning based
  diagnosis? the impact of domain shift in chest radiograph classification. In:
  {MICCAI} workshop on {Thoracic} {Image} {Analysis}. pp. 74--83. Springer
  (2020)

\bibitem{Camacho2021}
Quiñones‐Camacho, L.E., Fishburn, F.A., Belardi, K., Williams, D.L.,
  Huppert, T.J., Perlman, S.B.: Dysfunction in interpersonal neural
  synchronization as a mechanism for social impairment in autism spectrum
  disorder. Autism Research  \textbf{14}(8),  1585--1596 (2021)

\bibitem{Reindl2022}
Reindl, V., Wass, S., Leong, V., Scharke, W., Wistuba, S., Wirth, C.L., Konrad,
  K., Gerloff, C.: Multimodal hyperscanning reveals that synchrony of body and
  mind are distinct in mother-child dyads. NeuroImage  \textbf{251},  118982
  (May 2022)

\bibitem{Rozemberczki2020}
Rozemberczki, B., Sarkar, R.: Characteristic {Functions} on {Graphs}: {Birds}
  of a {Feather}, from {Statistical} {Descriptors} to {Parametric} {Models}.
  In: Proceedings of the 29th {ACM} {International} {Conference} on
  {Information} \& {Knowledge} {Management}. pp. 1325--1334. ACM, Virtual Event
  Ireland (Oct 2020)

\bibitem{Ruther2021}
Schulte‐Rüther, M., Kulvicius, T., Stroth, S., Wolff, N., Roessner, V.,
  Marschik, P.B., Kamp‐Becker, I., Poustka, L.: Using machine learning to
  improve diagnostic assessment of {ASD} in the light of specific differential
  and co‐occurring diagnoses. Journal of Child Psychology and Psychiatry
  (2022)

\bibitem{Scott2010}
Scott‐Van~Zeeland, A.A., Dapretto, M., Ghahremani, D.G., Poldrack, R.A.,
  Bookheimer, S.Y.: Reward processing in autism. Autism research
  \textbf{3}(2),  53--67 (2010)

\bibitem{Tanabe2012}
Tanabe, H.C., Kosaka, H., Saito, D.N., Koike, T., Hayashi, M.J., Izuma, K.,
  Komeda, H., Ishitobi, M., Omori, M., Munesue, T.: Hard to “tune in”:
  neural mechanisms of live face-to-face interaction with high-functioning
  autistic spectrum disorder. Frontiers in Human Neuroscience  \textbf{6}, ~268
  (2012)

\bibitem{Traut2022}
Traut, N., Heuer, K., Lemaître, G., Beggiato, A., Germanaud, D., Elmaleh, M.,
  Bethegnies, A., Bonnasse-Gahot, L., Cai, W., Chambon, S.: Insights from an
  autism imaging biomarker challenge: promises and threats to biomarker
  discovery. NeuroImage  \textbf{255},  119171 (2022)

\bibitem{Varoquaux2022}
Varoquaux, G., Cheplygina, V.: Machine learning for medical imaging:
  methodological failures and recommendations for the future. NPJ digital
  medicine  \textbf{5}(1), ~1--8 (2022)

\bibitem{Wang2021}
Wang, L., Huang, C., Ma, W., Cao, X., Vosoughi, S.: Graph {Embedding} via
  {Diffusion}-{Wavelets}-{Based} {Node} {Feature} {Distribution}
  {Characterization}. In: Proceedings of the 30th {ACM} {International}
  {Conference} on {Information} \& {Knowledge} {Management}. pp. 3478--3482.
  ACM, Queensland Australia (Oct 2021)

\bibitem{Wang2020}
Wang, Q., Han, Z., Hu, X., Feng, S., Wang, H., Liu, T., Yi, L.: Autism symptoms
  modulate interpersonal neural synchronization in children with autism
  spectrum disorder in cooperative interactions. Brain Topography
  \textbf{33}(1),  112--122 (2020)

\bibitem{Zeidan2022}
Zeidan, J., Fombonne, E., Scorah, J., Ibrahim, A., Durkin, M.S., Saxena, S.,
  Yusuf, A., Shih, A., Elsabbagh, M.: Global prevalence of autism: a systematic
  review update. Autism Research  \textbf{15}(5),  778--790 (2022)

\end{thebibliography}

\end{document}